\documentclass[journal]{IEEEtran}
\usepackage{amsmath,amsfonts,bm}

\def\eqref#1{equation~\ref{#1}}

\def\floor#1{\lfloor #1 \rfloor}
\def\1{\bm{1}}

\DeclareMathAlphabet{\mathsfit}{\encodingdefault}{\sfdefault}{m}{sl}
\SetMathAlphabet{\mathsfit}{bold}{\encodingdefault}{\sfdefault}{bx}{n}

\usepackage{graphicx}
\usepackage{url}
\usepackage{hyperref}
\usepackage{booktabs}
\usepackage{graphicx}
\usepackage{multirow}

\usepackage{floatrow}
\usepackage{makecell}

\usepackage{xcolor}

\ifCLASSINFOpdf
\else
\fi
\begin{document}
%
\title{SODAR: Exploring locally aggregated learning of mask representations for instance segmentation}
\title{SODAR: Segmenting Objects with Dynamically Aggregated Local Mask Representations}
\title{SODAR: Segmenting Objects by Dynamically Aggregating Local Mask Representations}
\title{SODAR: Segmenting Objects by Dynamically Aggregating Neighboring Mask Representations}
%
%
%

\author{
	Tao~Wang,
	Jun~Hao~Liew,
	Yu~Li,
	Yunpeng~Chen,
	Jiashi~Feng
	\thanks{Tao Wang is with NUS Graduate School of Integrative Sciences and Engineering and Institute of Data Science,  National University of Singapore, Singapore. E-mail: taowang@u.nus.edu.}
	\thanks{Jun~Hao~Liew and Jiashi~Feng are with the Department of Electrical and Computer Engineering, National University of Singapore, Singapore. E-mail: liewjunhao@u.nus.edu, elefjia@nus.edu.sg.}
	\thanks{Yu~Li is with Institute of Computing Technology, Chinese Academy of Science. E-mail: liyu@ict.ac.cn.}
	\thanks{Yunpeng~Chen is with YITU Technology, Beijing 100086, China. E-mail: yunpeng.chen@yitu-inc.com.}
	
}

\markboth{Journal of \LaTeX\ Class Files,~Vol.~14, No.~8, August~2015}%
{Shell \MakeLowercase{\textit{et al.}}: Bared Demo of IEEEtran.cls for IEEE Journals}

%



\maketitle

\graphicspath{{figures/}}

\begin{abstract}

Recent state-of-the-art one-stage instance segmentation model SOLO divides the input image into a grid and directly predicts  per grid  cell  object  masks  with  fully-convolutional  networks, yielding comparably good performance as traditional two-stage Mask R-CNN yet enjoying much simpler architecture and higher efficiency. We observe SOLO generates similar masks for an object at nearby grid cells, and these neighboring predictions can complement each other as some may better segment certain object  part,  most  of  which  are  however  directly  discarded  by non-maximum-suppression.  Motivated  by  the  observed  gap,  we develop a novel learning-based aggregation method that improves upon  SOLO  by  leveraging the rich neighboring  information while maintaining the architectural efficiency. The  resulting  model is named  SODAR. Unlike  the  original per  grid  cell  object  masks,  SODAR  is  implicitly  supervised  to learn mask representations that  encode  geometric  structure  of nearby  objects  and  complement  adjacent  representations  with context. The  aggregation  method  further  includes  two  novel designs:  1)  a  mask  interpolation  mechanism  that  enables  the model to generate much fewer mask representations by sharing neighboring representations among nearby grid cells, and thus  saves  computation  and  memory;  2)  a  deformable  neighbour sampling  mechanism  that  allows  the  model  to  adaptively  adjust neighbor sampling  locations  thus  gathering  mask representations with more relevant context and achieving higher performance. SODAR significantly  improves  the  instance segmentation performance, \textit{e.g.}, it  outperforms  a  SOLO  model  with  ResNet-101 backbone  by  2.2  AP  on  COCO  \texttt{test}  set,  with  only  about  3\% additional  computation.  We further show consistent performance gain with the SOLOv2 model. Code:\url{https://github.com/advdfacd/AggMask}
\end{abstract}

\begin{IEEEkeywords}
	Instance Segmentation, Object Detection, One-Stage, Feature Aggregation, Mask Representation
\end{IEEEkeywords}

\ifCLASSOPTIONpeerreview
\begin{center} \bfseries EDICS Category: 3-BBND \end{center}
\fi
%
\IEEEpeerreviewmaketitle

\begin{figure*}[!t]
	\centering
	\includegraphics[width=0.8\linewidth]{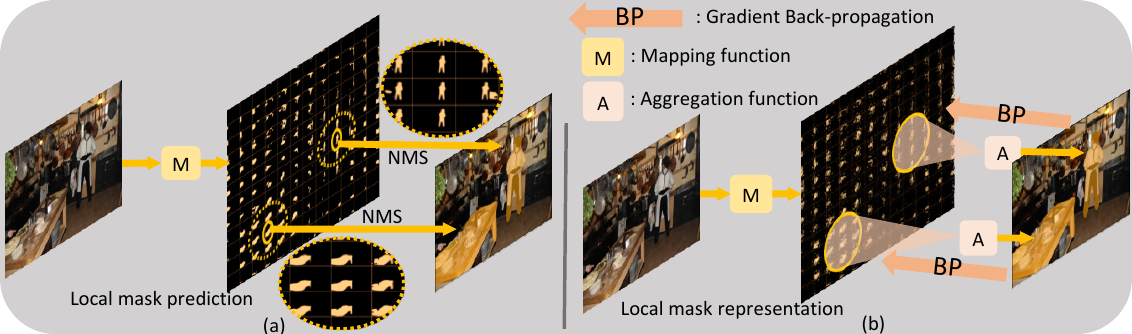}
	\caption{Illustration of our main idea. (a) State-of-the-art one-stage instance segmentation methods SOLO, SOLOv2 \cite{wang2019solo,wang2020solov2} evenly divide the input into grid cells and learn a mapping function with a fully convolution network for predicting per gird cell masks; the predictions at neighboring grid cells tend to be similar and complementary but do not contribute to the final result. (b) The proposed SODAR aims to exploit rich neighboring predictions by aggregating them in a learning-based fashion. 
	Meanwhile, by back-propagating through the aggregation function, the mapping function instead learns to generate local shape descriptors (mask representations) that encode shape and layout information of nearby objects and complement each other in combination, so as to generate masks of higher quality.}
	\label{intuition}
\end{figure*}

\section{Introduction}
\IEEEPARstart{S}{emantic} instance segmentation \cite{hariharan2014simultaneous}  aims to recognize and segment all the individual object instances of interested categories in an input image. 
Compared with standard object detection where an object is represented by a crude bounding box, instance segmentation requires an accurate pixel-wise binary mask that perfectly delineates the instance shape, which is thus more challenging.
Previously, detect-and-segment methods~\cite{he2017mask,liu2018path,chen2019hybrid,lee2019centermask,fu2019retinamask,chen2018masklab} dominate its solution, where objects are first localized using a bounding-box detection algorithm, followed by individual segmentation on each detected box.
Most of these approaches are inevitably slow due to the cumbersome box detection step and subsequent per region processing, and thus are hindered in real-time application and deployment on edge devices. Recent work of CenterMask~\cite{lee2019centermask} achieves excellent performance while maintaining high efficiency by adopting more efficient object detection framework~\cite{Tian2019}, lightweight backbone network~\cite{lee2019energy} and an attention-based mask prediction head, which helps focus on informative pixels for segmentation.
Though effective, the bounding-box representation often contains much background area and cannot distinguish highly overlapped objects with similar appearances~\cite{sofiiuk2019adaptis,shao2018crowdhuman}. And these methods heavily depend on accurate bounding-box detection as they cannot recover from errors in the object localization stage~\cite{sofiiuk2019adaptis,de2017semantic}, such as too small or shifted boxes.

In recent years, the community is focusing more on one-stage instance segmentation methods~\cite{chen2019tensormask,bolya2019yolact,xie2019polarmask,wang2019solo,wang2020solov2}, which simplify the detect-and-segment pipeline by directly predicting the object category and segmentation mask for each object instance. 
More recently, Wang \textit{et al.}~\cite{wang2019solo} proposed SOLO that evenly divides the input image into a spatial grid and predicts an object mask segmentation and classification score for each grid cell with fully convolutional networks~\cite{long2015fully}, achieving comparable results to the state-of-the-art. 
Compared with detect-and-segment methods~\cite{he2017mask,li2017fully},  the SOLO framework is much simpler by getting rid of bounding box detection and feature re-pooling. 
Its improved version SOLOv2~\cite{wang2020solov2} utilizes dynamic convolution to improve mask segmentation quality and further boosts model efficiency. The method decouples mask prediction into shared mask feature learning and dynamic convolution kernel prediction.

In this paper, we also try to improve the SOLO framework but from an orthogonal perspective with SOLOv2.
We observe that, in the SOLO framework, the mask predictions from dense per-grid cell mask segmentation are not fully exploited.
As shown in Fig.~\ref{intuition} (a), SOLO divides the input space spatially into a grid and directly predicts an object mask for each grid cell, followed by non-maximum suppression over all grid results to produce the final segmentation.
The mask predictions over a grid cell and its neighbors likely correspond to the same object instance.
They can complement each other as some may better segment certain object parts.
However, most of these predictions are directly discarded in non-maximum suppression, leading to loss of useful mask information.
We then naturally consider leveraging the complementarity of these neighboring predictions to further improve performance while maintaining the simple and efficient architecture design. 

A straightforward idea is to perform mask voting that averages the neighboring predictions that have sufficient overlap with the final result.
Similar ideas have been explored in the object detection field, known as box voting~\cite{gidaris2015object}.
In this work, we first carefully examine applying such a voting algorithm (see Section~\ref{pilotexp})
and find this approach indeed brings some improvement which however is very minor.
We hypothesize that unlike the bounding boxes, object masks are much more complicated due to deformations and layering, and thus cannot benefit much from simple post-training voting.

To better harness the rich neighboring predictions, we propose to adaptively combine them in a learning-based fashion. 
Specifically, we develop a learnable aggregation function that dynamically aggregates neighboring mask information for each grid to enrich its final mask prediction, as shown in Fig.~\ref{intuition} (b).
With gradients back-propagated through the aggregation function, the model learns to generate intermediate shape descriptors that capture the information of objects at nearby grid cells.
We name these shape descriptors \textit{mask representations}, which stand in contrast with original per grid cell object masks. 
The aggregation function can then gather complementary information from neighboring representations to improve segmentation quality.
To enable adaptive combination of neighboring mask representations, we instantiate the aggregation function with a small multi-layer convolution network with predicted weights.
This is motivated by the dynamic local filtering network~\cite{jia2016dynamic} that generates filter weights for adaptive local spatial transformation.
Such dynamic aggregation better handles the variations of object shape like diverse scales and spatial layouts.
Note that some recent works~\cite{wang2020solov2,tian2020conditional} also use dynamically instantiated networks for generating segmentation masks with a shared base feature. 
Compared with them, our dynamic aggregation approach is targeted at better combining neighboring information for final segmentation; actually our method can be combined with them.

The per grid cell masks are redundant and are costly in computation and memory.
This is also another intrinsic limitation of vanilla SOLO, which limits the spatial grid resolution.
As the proposed aggregated learning can generate mask representations that encode neighboring object information, we then  alleviate the spatial redundancy by allowing nearby grid cells to share the same set of representations and generate differentiated masks for them through dynamic aggregation.
This approach behaves similarly to bilinear interpolation and thus is named \textit{mask interpolation}.

A simple aggregation function is to employ a fixed neighbor sampling strategy, \textit{e.g.}, taking the spatially left and right neighbor.
However, natural objects have diverse scales and spatial layouts, and thus such setting is not optimal for covering as much relevant context information as possible. 
To enable more flexible neighbor selection, we develop a \textit{deformable neighbor sampling} mechanism to adaptively adjust the neighbor grid cell sampling locations. 
It is motivated by the deformable convolution operation~\cite{dai2017deformable}, where   offset are learnt so as to select proper neighboring mask representations.
With such a mechanism, the dynamic aggregation function is endowed with another dimension of freedom to adjust the pre-defined neighboring grid cell locations for higher performance.

The aggregation function, together with the proposed mask interpolation technique and deformable neighbor sampling mechanism, form a new instance segmentation model that first densely generates per grid cell mask representations and then adaptively selects and aggregates local neighboring representations to generate final segmentation. 
We name the model as SODAR, \textit{i.e.}, \textbf{S}egmenting \textbf{O}bjects by \textbf{D}ynamically \textbf{A}ggregating neighboring mask \textbf{R}epresentations.

To summarize, our contributions are as follows: 
\begin{itemize}
	\item We identify the discarded neighboring prediction issue of state-of-the-art one-stage instance segmentation models and carefully examine the widely used voting algorithm to leverage the neighboring predictions. We are the first to explore this problem.
	\item We develop a learning-based aggregation function, which enables the model to learn intermediate mask representations that capture context object information and improves final results by adaptively combining the neighboring representations.
	\item We further propose a mask interpolation mechanism and a deformable neighbor sampling mechanism. The former effectively reduces the number of generated mask representations to save computation and memory, and the latter enables the model to adjust the neighbor grid cell sampling locations for better performance.
	\item We conduct extensive experiments and model analysis on the COCO and high-quality LVIS benchmarks to validate the proposed method. On COCO, a ResNet-101 based SOLO model gets about 2.2 AP boost with only 3\% additional FLOPs. We also evaluate its generalization on SOLOv2~\cite{wang2020solov2}, TensorMask~\cite{chen2019tensormask} and CondInst~\cite{tian2020conditional}. 
	We believe the simple and effective framework and interpretable mask representations would shed light on future research.
\end{itemize}

\section{Related work}
We first review current instance segmentation methods, and then summarize existing approaches in many pattern recognition and computer vision fields that leverage neighboring information. 
Finally we revisit existing dynamically instantiated network designs.

\subsection{Instance Segmentation}
Instance segmentation~\cite{hariharan2014simultaneous} is a challenging computer vision task that requires precise localization of object instances at pixel level.
In recent years the community has made great progress by applying deep neural networks~\cite{simonyan2014very,he2016deep}, and existing methods can be summarized to three groups as below.

\subsubsection{Detect-and-segment Approaches}
Instance segmentation is previously tackled mainly by detect-and-segment approaches, or region-based methods~\cite{li2017fully,he2017mask,dai2016instance,chen2019hybrid,lee2019centermask,chen2018masklab} which predict an instance mask over the box region for each detected object. 
For example, FCIS~\cite{li2017fully} utilizes a position sensitive score map scheme to predict instance segmentation masks for object proposals; 
Mask R-CNN employs an additional convolutional mask branch on Faster R-CNN~\cite{ren2015faster} to predict instance masks;
Cascaded Mask R-CNN~\cite{cai2018cascade} and HTC~\cite{chen2019hybrid} further extend the detect-and-segment approach to multi-stage which achieves significant performance gain.
Some recent works~\cite{fu2019retinamask,lee2019centermask} replace the detection network with a fast one-stage model and employ a lightweight head for mask prediction.
Nevertheless, it is counter-intuitive to build instance segmentation upon bounding-box object detection as the bounding-box representation often contains much background area and cannot distinguish highly overlapped objects with similar appearances~\cite{sofiiuk2019adaptis,shao2018crowdhuman}. 
Moreover, these methods heavily depend on accurate bounding-box detection since they cannot recover from errors in the object localization stage~\cite{sofiiuk2019adaptis,de2017semantic}, such as too small or shifted boxes.

\subsubsection{One-stage Approaches}
The feature re-pooling and subsequent per ROI processing operations in detect-and-segment approaches are costly, which hinders their application in real-time scenarios and on edge devices with low computation resources.
Recently there has been a surge of interest in one-stage methods~\cite{bolya2019yolact,sofiiuk2019adaptis,chen2019tensormask,tian2020conditional,xie2019polarmask,chen2020blendmask,qi2020pointins,zhang2020mask, wang2019solo, wang2020solov2}, which directly predict instance masks along with classification scores. 
For example, YOLACT~\cite{bolya2019yolact} employs a mask assembly mechanism that linearly combines prototypes with learned weights;
TensorMask~\cite{xie2019polarmask} generalizes dense object detection to dense instance segmentation with an aligned feature representation;
SOLO~\cite{wang2019solo} directly predicts per grid cell masks with a specifically designed FCN architecture and learning objectives, which is the current SOTA one-stage instance segmentation architecture; 
The most recent SOLOv2~\cite{wang2020solov2} improves SOLO by replacing the mask prediction head with dynamic convolution~\cite{jia2016dynamic}. 
In this work, we further boost the SOTA one-stage framework by designing a novel dynamic neighbor aggregation method.

\subsubsection{Bottom-up Approaches}
Bottom-up instance segmentation methods~\cite{de2017semantic,liu2017sgn,gao2019ssap,newell2017associative} group similar pixels to form the segmentation mask.
For example,~\cite{de2017semantic} designs a discriminative loss function to learn pixel representation such that the pixels can be clustered to produce instance segmentation;
Liu \textit{et al.}~\cite{liu2017sgn} introduce a sequential grouping framework, in which pixels are first grouped into lines that are then grouped into connected components, and finally the components are grouped to form the instance-level mask;
similar to~\cite{de2017semantic}, \cite{gao2019ssap} learns pixel affinity maps that are then fed to  a graph partition module to group pixels into segmentation masks. 
The bottom-up approaches are shown to be better at handling occlusion than detect-and-segment approaches in traffic scenes (such as the CityScapes dataset~\cite{cordts2016cityscapes}), 
but currently have much lower performance on common object detection datasets like COCO~\cite{lin2014coco} and Pascal VOC~\cite{everingham2010pascal}.

\subsection{Leveraging Neighboring Information}
Aggregating information from the neighborhood is a prevailing approach adopted in many learning methods and tasks. 
For example, the classical K nearest neighbor classification method~\cite{fix1951discriminatory} utilizes neighboring data points to vote for query data;
the convolution networks~\cite{fukushima1980neocognitron,lecun1995convolutional} can be viewed as stacking convolution operators that fuse information from spatially neighboring pixels to get new representation at each spatial location.
Besides, in point cloud processing,~\cite{Qi2017} proposes a new backbone design based on hierarchical neighboring data point aggregation and is able to significantly improve previous PointNet~\cite{qi2017pointnet}.
Graph nerual networks (GNN)~\cite{scarselli2008graph,battaglia2016interaction} including variants Graph convolutional networks (GCN)~\cite{kipf2016semi} and GraphSAGE~\cite{hamilton2017inductive} follow a recursive neighborhood aggregation and transformation scheme to learn high quality representation of graph data.  
In video action recognition, the motion information encoded in temporal neighboring frames has been shown crucial for good performance~\cite{Feichtenhofer, wang2017spatiotemporal}.
A work by He \textit{et al.}~\cite{He2019} is most relevant to our method, in which the bounding box regression uncertainty loss is learnt and the uncertainty predictions are  then used to vote for the final bounding box with neighboring boxes.
In this work, we explore the neighboring information aggregation over mask predictions to further boost one-stage instance segmentation.

\subsection{Dynamically Instantiated Networks}
Early neural networks employ fixed weights to process input signals, which are thus independent of the input and learned with gradient back-propagation.
In recent years researchers explore various dynamic network weight designs, which enable input-dependent weights and introduce more adaptability and flexibility to neural networks. 
For example,~\cite{jaderberg2015spatial} first introduces a dynamic parametric transformation to adaptively transform the intermediate feature map, with controlling parameters conditioned on the input.
Remarkably, \cite{jia2016dynamic} proposes a dynamic filter network with parameters predicted with the input to more flexibly process the input signal.
Deformable convolution~\cite{dai2017deformable} can also be viewed as a dynamically instantiated network design where the regular sampling location of convolution operation is improved to dynamic location conditioned on the input.
Dynamic networks have been recently employed to improve instance segmentation models. 
For example, YOLACT~\cite{bolya2019yolact} generates mask coefficients to combine a set of corresponding mask prototypes to generate the segmentation mask.
BlendMask~\cite{chen2020blendmask} extends the scalar coefficients in YOLACT to two dimensional coefficients and achieves better results with less mask prototypes.
AdaptIS~\cite{sofiiuk2019adaptis} employs adaptive instance normalization to generate the segmentation mask, with its affine parameters conditioned on the input.
CondInst~\cite{tian2020conditional} and SOLOv2~\cite{wang2020solov2} apply dynamic convolution to generate instance masks based on globally shared feature maps. 
In this work, we also utilize a dynamic network design.
We specifically compare our method with most relevant YOLACT, BlendMask, CondInst, SOLOv2 and explain the differences. Tab.~\ref{compare_baseline} gives a summary.
YOLACT and BlendMask use normalized coefficients to linearly combine base mask features and require box detection to eliminate noise outside the object area.
Comparatively, our SODAR employs the general dynamic convolution and does not require box detection.
Moreover, the mask features for previous methods are globally shared for all spatial locations while our proposed method learns mask representations that are only locally shared. 
CondInst and SOLOv2 do not need box detection and also use dynamic convolution, but the base features are also globally shared as YOLACT and BlendMask. 
More importantly, our method aims at learning a dynamic network for leveraging neighboring predictions and is orthogonal to CondInst and SOLOv2. 
We show in experiments that the SOLOv2 model is effectively improved when augmented with our proposed aggregation function.

\begin{table}[t]
	\newcommand{\tabincell}[2]{\begin{tabular}{@{}#1@{}}#2\end{tabular}}  
	\renewcommand{\tabcolsep}{1pt}
	\renewcommand{\arraystretch}{1.1}
	\begin{tabular}{c|ccc}
		-&  \tabincell{c}{YOLACT~\cite{bolya2019yolact}/\\BlendMask~\cite{chen2020blendmask}} & \tabincell{c}{SOLO~\cite{wang2020solov2}/\\SOLOv2~\cite{wang2020solov2}}& SODAR \\ \hline
		base feature & globally-shared & globally-shared & locally-shared\\
		box detection & \checkmark & - & - \\
		dynamic conv & -& \checkmark & \checkmark\\
		neighbor info & -& - & \checkmark
	\end{tabular}
	\caption{Comparison of model details with most relevant instance segmentation models. ``Base feature'' means the base feature used for generating the final mask segmentation; ``box detection'' and ``dynamic convolution'' mean whether or not box detection is required or dynamic convolution is used; ``neighbor info'' denotes utilizing neighboring information.}
	\label{compare_baseline}
\end{table}


\section{Preliminaries}
\label{preliminaries}
In this section, we revisit the general architecture and formulation of SOLO and present pilot experiments on applying the mask voting algorithm to leveraging neighboring mask information.

\subsection{SOLO Formulation}
For an input image $I$, instance segmentation aims to predict a set of $\{c_{k},M_{k}\}_{k=1}^n$, where $c_{k}$ is the object class label, $M_{k}$ is a binary 2-d map that shows segmentation of the object, and $n$ varies with the number of objects in the image.
Without loss of generality, SOLO~\cite{wang2019solo} spatially divides the input image into a $G \times G$ grid and learns two simultaneous mapping functions for predicting the semantic category $c_{ij}$ and the corresponding mask segmentation $M_{ij}$ for each spatial grid cell:
\begin{equation}
\mathcal{F}_{c}(I, \theta_{c}): I \mapsto \{c_{ij}\in\mathbb{R}^{C} |i,j=0,1,\ldots,G\},
\label{mapping_cls}
\end{equation}
\begin{equation}
\mathcal{F}_{m}(I, \theta_{m}): I \mapsto \{M_{ij}\in\mathbb{R}^{H \times W}|i,j=0,1,\ldots,G\}.
\label{mapping_mask_representation}
\end{equation}
Here $\theta_{c}$ and $\theta_{{m}}$ denote parameters for the above two mapping functions respectively. 
$\mathcal{F}_{c}$ and $\mathcal{F}_{m}$ are implemented with fully convolutional networks and share the backbone parameters.
$C$ is the number of object categories. 
Each element of $c_{ij}$ is within the range $[0,1]$ and indicates whether an object of the corresponding category exists at the grid cell $(i,j)$.
$H$, $W$ are the height and width of the output mask that aligns with the input image content. 
With the semantic category and mask segmentation predicted for each spatial grid cell, the instance segmentation result is acquired by simple post-processing of non-maximum suppression (NMS).
SOLOv2~\cite{wang2020solov2} improves SOLO by replacing the mask prediction head with dynamic convolution but the general instance segmentation architecture remains the same.
For a detected object at the grid cell $(i,j)$, the model tends to generate similar mask predictions at nearby grid cells, \textit{e.g.,} $M_{i,j-1}$, but this $M_{i,j-1}$ is discarded by NMS and does not contribute to the final prediction.

\begin{table}[t!]
	\renewcommand{\tabcolsep}{3.5pt}
	\renewcommand{\arraystretch}{1.1}
	\begin{tabular}{c|c|c|c|c}
		- &  baseline & +avg.  v.& +score v. & +IOU v. \\ \hline
		AP & 35.8 & 36.0 & 36.1 & 36.1\\
	\end{tabular}
	\caption{Results of different voting strategies with SOLO-ResNet50. ``v.'' stands for voting.}%
	\label{voting_result}
\end{table}

\begin{figure}[t!]
	\includegraphics[width=0.8\linewidth]{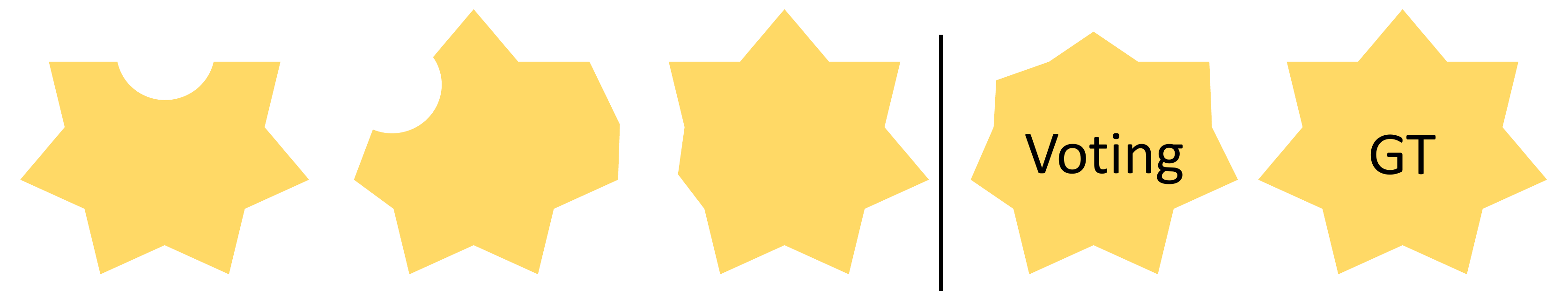}
	\caption{Mask voting example. The 3 masks on the left of the black line are the base predictions.
		The 2 masks on the right are voting results (Voting) and the desired ground truth (GT). Voting generates a sub-optimal mask.}%
	\label{toy_voting}
\end{figure}

\begin{figure*}[t!]
	\centering
	\includegraphics[width=0.8\linewidth]{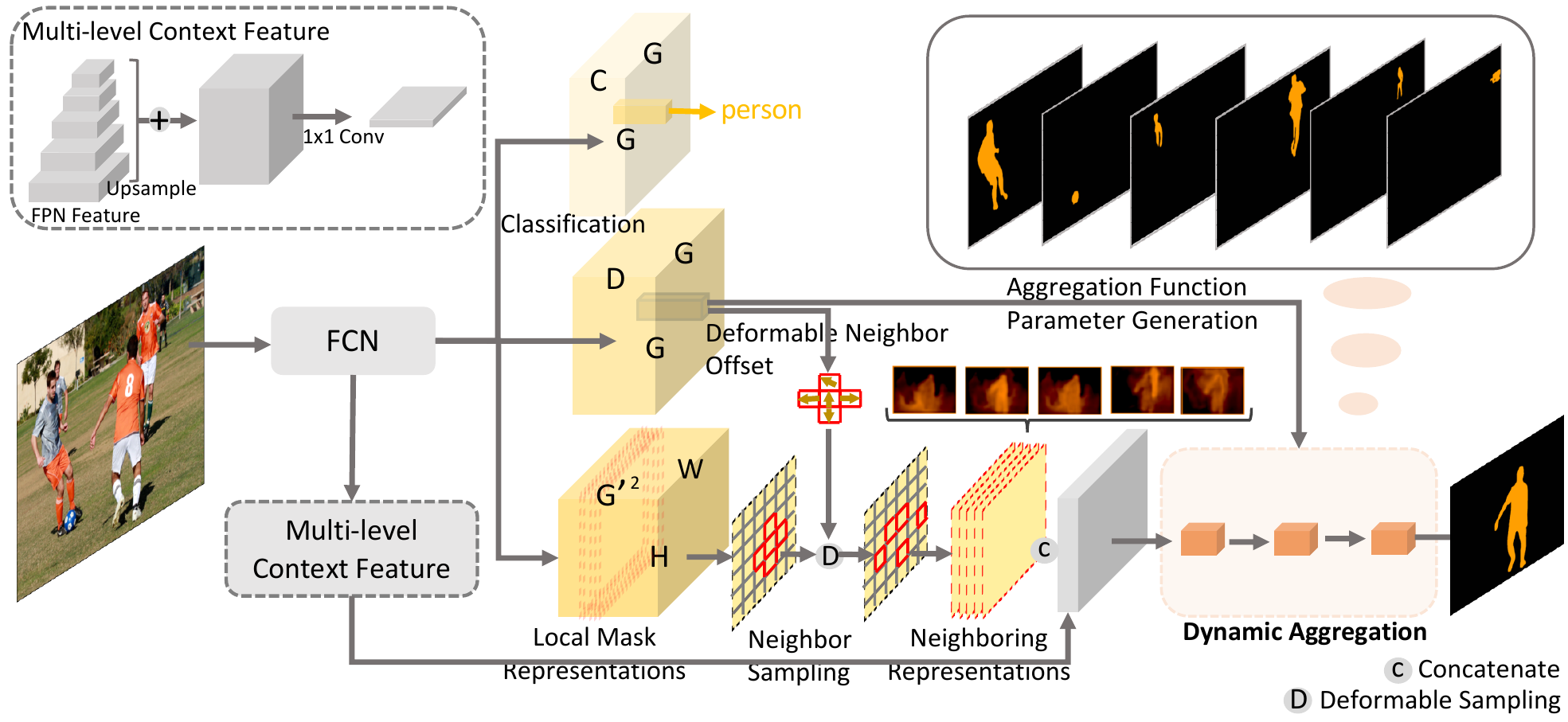}
	\caption{Illustration of the network architecture. For objects at different grid cells, their corresponding spatially neighboring mask representations are gathered and aggregated to form the segmentation. 
	The deformable neighbor sampling mechanism allows the model to adaptively adjust the neighbor sampling locations to gather mask representations with relevant context information. 
	The framework is fully convolutional and end-to-end trained. 
	Note with mask interpolation, the grid resolution of mask representation ($G^{\prime}$) can be smaller than classification ($G$) to save computation and memory. }
	\label{net_arch}
\end{figure*}


\subsection{Pilot Experiments: Mask Voting}
\label{pilotexp}
In this subsection, we present pilot experiments and analysis to check whether the SOLO model can benefit from rich neighboring predictions by adopting the voting algorithm~\cite{gidaris2015object}.
We compute mask IOU as the similarity measure to gather neighboring mask predictions used for voting with three schemes: 
1) simple averaging (\textbf{avg. v.}); 2) weighted averaging with corresponding classification scores (\textbf{score v.}); 3) weighted averaging with corresponding IOU (\textbf{IOU. v.}). 
For each scheme, we perform grid search to find the optimal IOU threshold used for voting. 
As shown in Tab.~\ref{voting_result}, refining the mask prediction during inference with the voting procedure can indeed improve the final instance segmentation performance (\textit{e.g.,} 0.2 AP with simple averaged voting). 
However, even with weighted voting, the improvement is minor and at most 0.3 in AP. 
The voting procedure is essentially simple post-processing and requires no modification to a trained model, but an object shape may have complex spatial layout, and thus the simple voting algorithm may be suboptimal. 
Fig.~\ref{toy_voting} gives a toy example to illustrate that voting may provide suboptimal results as it averages the candidates and is incapable of selectively combining the segmentation of different parts.

\section{Proposed Method}
To more effectively leverage the neighboring predictions, we propose an aggregated learning approach, with which we build the instance segmentation model named SODAR, \textit{i.e.,} \textbf{S}egmenting \textbf{O}bjects by \textbf{D}ynamically \textbf{A}ggregating neighboring mask \textbf{R}epresentations.
Specifically, we aggregate the neighboring mask predictions with a learnable aggregation function:
\begin{equation}
\dot{M}_{ij}= \mathrm{Agg}(M_{ij} \oplus \mathcal{N}(M_{ij}),\theta_{a}).
\label{eq_aggregation1}
\end{equation}
Here $\mathrm{Agg}(\cdot)$ and $\theta_{a}$ denotes aggregation function and its parameter respectively.
$\oplus$ means the concatenation operation, $\dot{M}$ denotes the final aggregated mask prediction and $\mathcal{N}(M_{ij})$ refers to the set of neighboring predictions that will be used for predicting the final instance mask $\dot{M}_{ij}$ at grid cell $(i,j)$. 
We consider using either 4 or 8 nearest neighbors, \textit{i.e.}, $\{M_{i-1,j}, M_{i,j-1}, M_{i+1,j}, M_{i,j+1}\}$ or $\{M_{i+p,j+q}\} \backslash M_{ij}$ with $p,q\in \{-1,0,1\}$. 
More complex designs may further improve performance but are beyond the scope of this work.

The mask prediction $M_{ij}$ is forced to be within the value range $(0,1)$ with the sigmoid activation function.
To allow more freedom for learning the intermediate mask representation, we remove the sigmoid activation before aggregation.
To complement the top-down instance specific information in mask representation, we append a bottom-up multi-level context feature $F_c$ of only a few channels that summarizes both high-level semantics and low-level fine details to guide the aggregation process:
\begin{equation}
\dot{M}_{ij}= \mathrm{Agg}(\hat{M}_{ij}  \oplus \mathcal{N}(\hat{M}_{ij}) \oplus F_c,\theta_{a}),
\label{eq_aggregation2}
\end{equation} 
where $\hat{M}_{ij}$ is raw mask prediction without sigmoid activation, \textit{i.e.,} the mask representations to be learned. The aggregation is differentiable and no explicit supervision is imposed for learning $\hat{M}_{ij}$.

\paragraph{Dynamic Aggregation} In Eqn. (\ref{eq_aggregation2}), the same set of aggregation parameters $\theta_{a}$ is shared across all grid cells.
However, such an object-invariant aggregation scheme may be suboptimal for desired adaptive neighboring mask combination as object shape varies significantly with different spatial layouts, scales and context.
Inspired by dynamic local filter proposed in~\cite{jia2016dynamic}, we extend our aggregation function by additionally learning a mapping function that generates position-specific aggregation parameters:
\begin{equation}
\mathcal{F}_{a}(I, \theta_{a}): I \mapsto \{\theta_{ij}\in\mathbb{R}^{D}|i,j=0,1,\ldots,G\}.
\label{eq_aggregation_param_generate}
\end{equation}
Here $I$ denotes input image, $D$ is the dimension of aggregation network parameters, in implementation, we reshape it to the corresponding convolution kernels of the aggregation network. We then apply the predicted parameters for aggregating neighboring local mask representations:
\begin{equation}
\dot{M}_{ij}= \mathrm{Agg}(\hat{M}_{ij} \oplus \mathcal{N}(\hat{M}_{ij}) \oplus F_c, \theta_{ij}).
\label{eq_dynamic_aggregation}
\end{equation}
Such a dynamic aggregation scheme not only achieves desired self-adaptive combination of neighboring mask predictions, but also enables an interesting extension of SODAR, which is discussed next.

\begin{figure}[t!]
	\centering
	\includegraphics[width=0.8\linewidth]{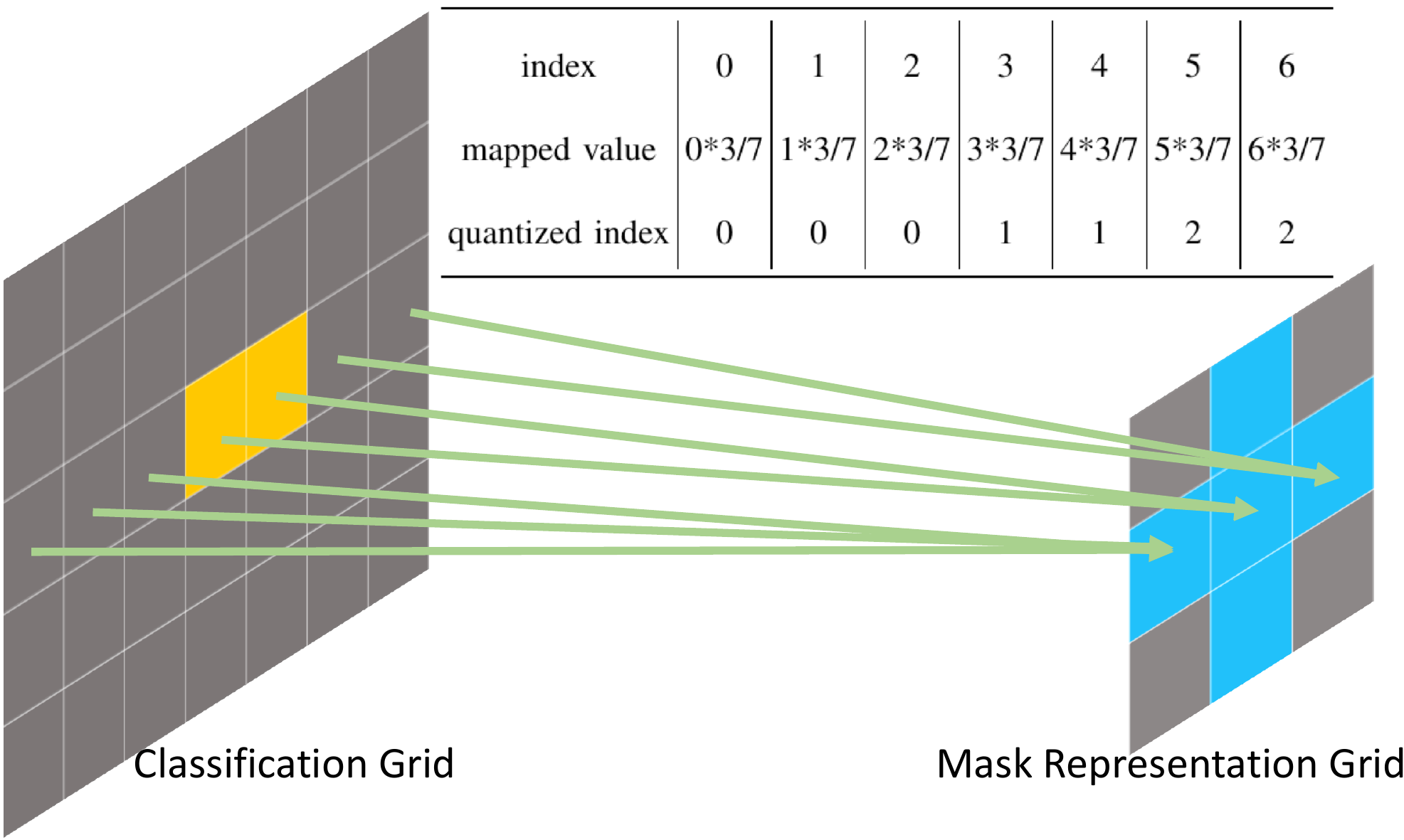}
	\caption{Illustration of the mask interpolation mapping mechanism. The example maps a  5x7 classification grid to a 3x3 mask representation grid. The two grid cells (orange) in the left grid are mapped to the same grid cell in the right gird and share the 5 neighboring mask representations (blue). The table shows the index mapping calculation.}
	\label{mask-itp}
\end{figure}

\paragraph{Mask Interpolation}
In the SOLO model, the grid resolution is limited as finer grids would result in increasingly many mask representations and cause remarkably increased computation and memory cost, \textit{e.g.}, a $20\times20$ grid leading to 400 mask predictions, while $40\times40$ leading to 1,600 predictions. 
However, a finer grid, especially on low-level features, is beneficial to recognizing the small objects or distinguishing the nearby objects. 
To address the limitation, we extend SODAR to a variant with a larger grid resolution $G$ for classification but a smaller grid resolution $G^{\prime}$ ($G^{\prime}<G$) for mask representation. 
The mapping functions for mask representation then becomes:
\begin{equation}
\mathcal{F}_{\hat{m}}(I, \theta_{\hat{m}}): I \mapsto \{\hat{M}_{ij}\in\mathbb{R}^{H \times W}|i,j=0,1,\ldots,G^{\prime}\}.
\label{mapping_mask_representation_coarser}
\end{equation}
To generate mask predictions with finer classification grids, we simply allow nearby classification grid cells to share the same set of neighboring mask representations for predicting the object mask:
\begin{equation}
\dot{M}_{ij}= \mathrm{Agg}(\hat{M}_{\floor{i*\frac{G^{\prime}}{G}}\floor{j*\frac{G^{\prime}}{G}}} \oplus \mathcal{N}(\hat{M}_{\floor{i*\frac{G^{\prime}}{G}}\floor{j*\frac{G^{\prime}}{G}}}) \oplus F_c, \theta_{ij}).
\label{eq_mask_interpolation}
\end{equation}
For instance, when the grid resolution $G^{\prime}$$=$$10$ for mask representation and $G$$=$$20$ for classification, both the masks $\dot{M}_{15,15}$ and $\dot{M}_{14,14}$ are predicted using the same set of neighboring representations of $\hat{M}_{7,7}$, but with different dynamically generated weights $\theta_{15,15}$ and $\theta_{14,14}$. Fig.~\ref{mask-itp} gives an illustration of the mapping mechanism.
This can be viewed as a spatial interpolation operation to obtain finer grid mask predictions given coarser grid mask representations, which we name \textit{mask interpolation}. 
It effectively reduces the number of representations needed for a certain grid resolution, offering two advantages.
1) We can increase a model's classification grid resolution to improve its discrimination ability for scenes of more dense or small objects, with small computation overhead; 2) we can decrease the mask representation grid resolution to save computation and memory cost without sacrificing much performance.

\begin{figure}[h!]
	\centering
	\includegraphics[width=1\linewidth]{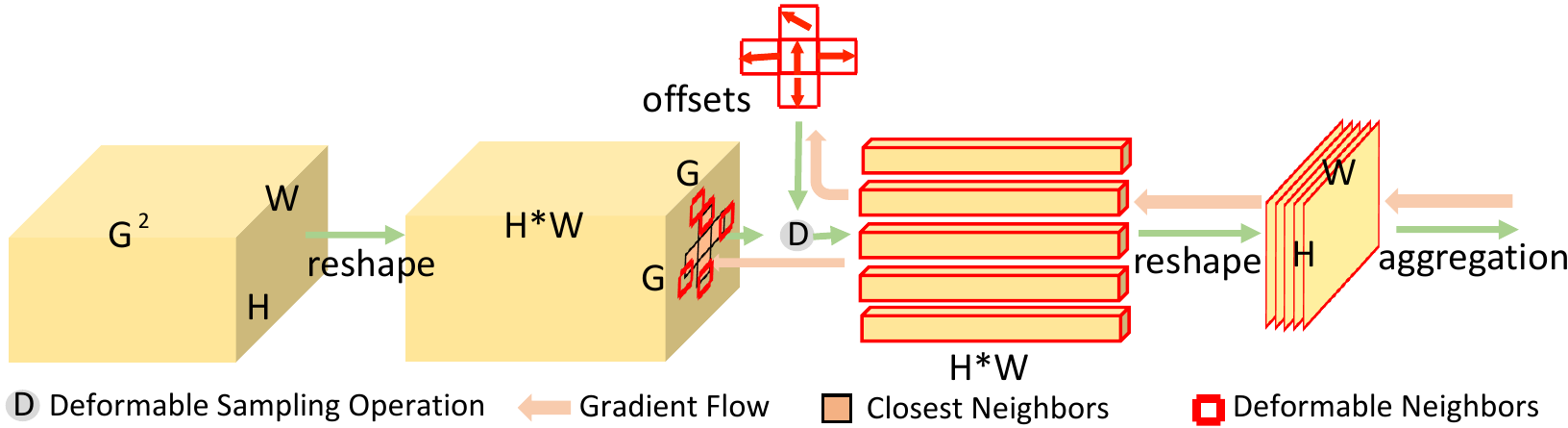}
	\caption{Illustration of deformable neighbor sampling implementation. 
		To learn the deformable offset prediction,
		we reshape the mask representations and view them as regular convolution feature maps to perform the deformable neighbor sampling as in deformable convolution. The gradient is back-propagated to learn the offset prediction.}
	\label{deform_neighbor}
\end{figure}

\paragraph{Deformable Neighbors}
In the above introduced aggregation framework,
the neighboring mask representations is the closest 4 or 8 spatial nearest neighbors, \textit{i.e.}, $\{\hat{M}_{i-1,j}, \hat{M}_{i,j-1}, \hat{M}_{i+1,j}, \hat{M}_{i,j+1}\}$ or $\{\hat{M}_{i+p,j+q}\} \backslash \hat{M}_{ij}$ with $p,q\in \{-1,0,1\}$.  
However, this setting may not be optimal as the neighbor grid cell locations are fixed. 
To enable more flexible neighboring representation selection, we introduce a deformable neighbor sampling mechanism. 
Motivated by the deformable convolution operation~\cite{dai2017deformable}, we additionally predict the spatial offset for each neighboring mask representation,
and then gather the representations at the offseted grid cells for latter aggregation.
The offset comprises two offset values accordingly in x and y axes. 
Formally, the offsets are predicted as
\begin{equation}
\mathcal{F}_{off}(I, \theta_{off}): I \mapsto \{\Delta_{ij}\in\mathbb{R}^{P}|i,j=0,1,\ldots,G\}.
\label{eq_offset_prediction}
\end{equation}
$\Delta_{ij}$ is a P-dimensional offset vector for all the gathered neighboring mask representations; \textit{e.g.}, for the 4-neighbor setting, there are 5 neighboring representations, \textit{i.e.}, the top, bottom, left, right and center representations. 
Then $P=10$ and $\Delta_{ij}$ contains the offsets for each of the gathered representations:
\begin{equation}
\begin{split}
\Delta_{ij} =&\{(\Delta py^{top}_{ij},\Delta px^{top}_{ij}),(\Delta py^{left}_{ij},\Delta px^{left}_{ij}),(\Delta py^{center}_{ij},\\ &\Delta px^{center}_{ij}),(\Delta py^{bottom}_{ij},\Delta px^{bottom}_{ij}),\\ &(\Delta py^{right}_{ij},\Delta px^{right}_{ij})\}.
\end{split}
\end{equation}
Denote the mask representations gathered at grid cell $(i,j)$ as $\{\hat{M}_{i-1,j}, \hat{M}_{i,j-1}, \hat{M}_{i,j}, \hat{M}_{i+1,j}, \hat{M}_{i,j+1}\}$. Then the offseted representations are:
\begin{equation}
\begin{split}
&\{\hat{M}_{i-1+\Delta py^{top}_{ij},j+\Delta px^{top}_{ij}}, \hat{M}_{i+\Delta py^{left}_{ij},j-1+\Delta px^{left}_{ij}}, \\ & \hat{M}_{i+\Delta py^{center}_{ij},j+\Delta px^{center}_{ij}}, \hat{M}_{i+1+\Delta py^{bottom}_{ij},j+\Delta px^{bottom}_{ij}},\\ & \hat{M}_{i+\Delta py^{right}_{ij},j+1+\Delta px^{right}_{ij}}\}.
\end{split}
\end{equation}
The offset is fractional, so bilinear interpolation is employed to obtain the offseted mask representation, as in~\cite{dai2017deformable}.
To implement deformable offset learning like deformable convolution, as shown in Fig.~\ref{deform_neighbor}, we transform the mask representations from shape $(H,W,G^2)$ to shape $(G,G,HW)$ by flattening the spatial dimension and reshaping the channel dimension.
The gradient from latter dynamic aggregation is back-propagated to learn the offset prediction. 
The offsets are predicted together with the dynamic aggregation weights by adding additional channels. 
The deformable neighbor sampling mechanism introduces more versatility to the dynamic aggregation design. 
It can be combined with the above mask interpolation, by replacing the indexing with the mapping as above, \textit{e.g.}, $\hat{M}_{\floor{i*\frac{G^{\prime}}{G}}+\Delta py^{center}_{ij}, \floor{j*\frac{G^{\prime}}{G}}+\Delta px^{center}_{ij}}$.

\vspace{1em}
\paragraph{Network Architecture}
Fig.~\ref{net_arch} depicts the architecture of SODAR. The mapping function $\mathcal{F}_{a}$ shares the feature extraction network with $\mathcal{F}_{c}$ and $\mathcal{F}_{m}$.
We instantiate the dynamic aggregation function with a small multi-layer convolution network, which learns arbitrary nonlinear transformation. 
The multi-level context information is obtained by upsampling and combining FPN~\cite{lin2017feature} features, followed by 1x1 convolution to obtain compact feature (\textit{e.g.}, 16 channels). 
The whole model is fully convolutional and trained with the same loss formulation as in SOLO~\cite{wang2020solov2}.

%

\begin{figure*}[h]
	\centering
	\includegraphics[width=0.85\linewidth]{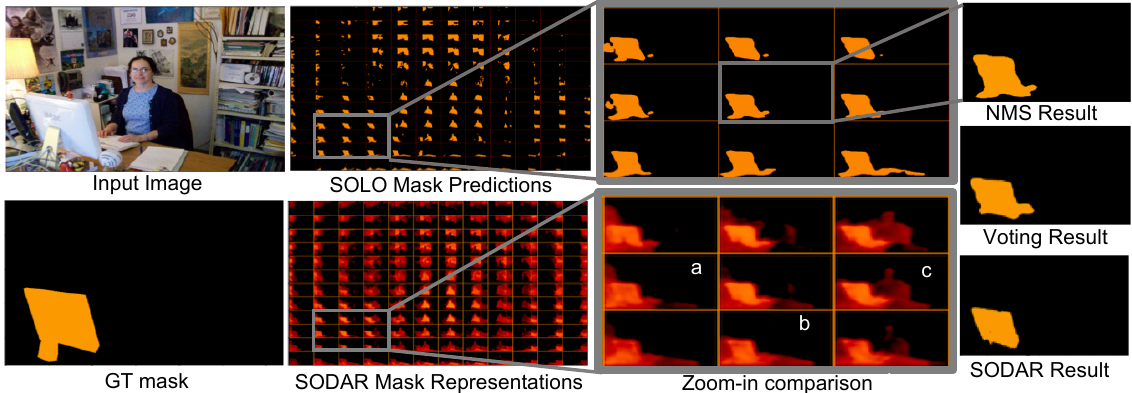}
	\caption{Visualization of mask representations learned by SODAR. Compared to baseline SOLO mask prediction, 
	our mask representation exhibit the following characteristics:
	1) has larger high-response area, indicating its capability to attend to the surrounding context;
	and 2) captures complementary information, \textit{e.g.}, in addition to the monitor, ``a'' attends to the lamp while ``b'' and ``c'' capture the desk and person, respectively.
	Thanks to this, the segmentation quality is higher than both the baseline model and simple voting algorithm. 
	Note most of the monitor stands in the training data are not labeled and treated as background. As a result, the model is not able to learn to segment the stand part from the data. Therefore, istead of computer stand, SOLO is more likely misguided by the color and appearance to segment the additional surrounding background area (with similar white color and apperance to the monitor). In contrast, SODAR learns from the data to only segment the screen part and avoided the misguiding background surrounded, which is reasonable.
	}
	\label{vis_mask_rep}
\end{figure*}\vspace{-1em}

\section{Experiments}

\subsection{Dataset}
We adopt COCO~\cite{lin2014coco} and LVIS~\cite{gupta2019} datasets for experiments. 
We report the standard COCO-style mask AP using the median of 3 runs. 
As AP for COCO may not fully reflect the improvement in mask quality due to its coarse ground-truth annotations~\cite{gupta2019,kirillov2020pointrend}, we additionally report AP evaluated on the 80 COCO category subset of LVIS, which has high-quality instance mask annotations, denoted as AP$^*$. AP$^*$ can better reveal the mask prediction quality, especially under a high IOU criteria.
Note we directly evaluate COCO-trained models against higher-quality LVIS annotations without training on it.

\subsection{Models and Implementation Details}
\begin{figure}[t!]
	\centering
	\includegraphics[width=0.99\linewidth]{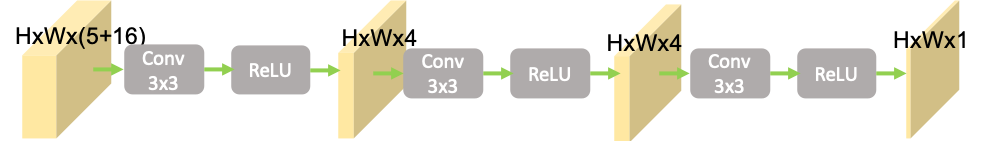}
	\caption{Illustration of the network architecture of the dynamic aggregation function. The input feature is composed of 5 channels of neighboring mask representations and 16 channels of the multi-level context feature, the intermediate feature has channel dimension of 4, the output is one channel of mask segmentation. Such a small multi-layer network is lightweight.}
	\label{agg_arch}
\end{figure}

In addition to SOLO~\cite{wang2019solo}, we also examine the generalizability of the proposed method to the SOLOv2~\cite{wang2020solov2} model.
Unless otherwise specified, we use 4 neighbors for aggregation and 16 channels for multi-level context feature.
We modify the existing deformable convolution code\footnote{\href{https://github.com/4uiiurz1/pytorch-deform-conv-v2}{\url{https://github.com/4uiiurz1/pytorch-deform-conv-v2}}} to implement the deformable sampling. As shown in Fig.~\ref{agg_arch}, by default, we instantiate the dynamic aggregation function with three consecutive layers of Convolution and ReLU operations.
To enable batched computation of different grid cells, we implement the dynamic aggregation with a 3-layer group convolution network, \textit{i.e.}, each batch item corresponds to one independent group. 
The grid resolution, except for the mask interpolation experiment, is set as 40, 36, 24, 16 and 12 for different FPN levels.
Other settings, including the training schedule, loss weight, and label assignment rules are the same as SOLO and SOLOv2 for fair comparison. 

\begin{table*}[h!]
	\centering
	\renewcommand{\tabcolsep}{2.0pt}
	\renewcommand{\arraystretch}{1.1}
	\begin{tabular}{c|cccccc|ccccccc}
		\toprule
		Model & AP & AP$_{50}$ & AP$_{75}$ & AP$_{s}$ & AP$_{m}$ & AP$_{l}$ & AP$^*$ & AP$^*_{50}$ & AP$^*_{60}$ & AP$^*_{70}$ & AP$^*_{80}$ & AP$^*_{90}$ & FPS \\\midrule
		SOLO-R50 & 35.8  & 57.1  & 37.8  & 15.0  & 37.8  & 53.6  & 37.0  & 58.2  & 51.8  & 43.6  & 32.2  & 14.2  & 12.9 \\
		SODAR-R50 & 37.9  & 58.2  & 40.9  & 16.2  & 41.5  & 56.3  & 40.0  & 60.4  & 54.9  & 47.1  & 36.4  & 17.4  & 11.4 \\
		$\Delta$-    & +2.1   & +1.1   & +3.1   & +1.2   & +3.7   & +2.7   & +3.0   & +2.2   & +3.1   & \textbf{+3.5}   & \textbf{+4.2}   & \textbf{+3.2}   & - \\ \midrule
		SOLO-R101& 37.1  & 58.1  & 39.5  & 15.6  & 41.0  & 55.5  & 38.6  & 59.7  & 53.5  & 45.8  & 34.3  & 15.6  & 11.5 \\
		SODAR-R101 & 38.6  & 58.9  & 41.7  & 16.5  & 42.6  & 57.7  & 41.2  & 61.9  & 55.8  & 48.4  & 37.7  & 18.5  & 10.2 \\
		$\Delta$-    & +1.5   & +0.8   & +2.2   & +0.9   & +1.6   & +2.2   & +2.6   & +2.2   & +2.3   & \textbf{+2.6}   & \textbf{+3.4}   & \textbf{+2.9}   & - \\ \midrule  \midrule
		SOLOv2-R50 & 37.7  & 58.5  & 40.2  & 15.6  & 41.3  & 56.6  & 40.0  & 60.3  & 54.4  & 47.0  & 36.5  & 18.5  & 16.4 \\
		SODAR*-R50 & 38.7  & 59.2  & 41.6  & 17.1  & 42.5  & 57.2  & 41.0  & 60.9  & 54.9  & 47.9  & 37.7  & 19.9  & 14.8 \\
		$\Delta$-    & +1.0   & +0.7   & +1.4   & +1.5   & +1.2   & +0.6   & +1.0   & +0.6   & +0.5   & \textbf{+0.9}   & \textbf{+1.2}   & \textbf{+1.4}   & - \\ \midrule
		SOLOv2-R101 & 38.5  & 59.1  & 41.3  & 17.1  & 42.5  & 56.8  & 41.1  & 61.9  & 55.9  & 48.0  & 37.3  & 18.8  & 13.4 \\
		SODAR*-R101 & 39.4  & 60.0  & 42.3  & 16.8  & 43.6  & 58.5  & 42.3  & 62.3  & 56.9  & 49.7  & 38.9  & 20.5  & 12.2 \\
		$\Delta$-    & +0.9   & +0.9   & +1.0   & -0.3  & +1.1   & +1.7   & +1.2   & +0.4   & +1.0   & \textbf{+1.7}   & \textbf{+1.6}   & \textbf{+1.7}   & - \\ 
		\bottomrule
	\end{tabular}
	\vspace{-1em}\caption{Comparison with baselines SOLO~\cite{wang2019solo} and SOLOv2~\cite{wang2020solov2}. SODAR* means SODAR model built on SOLOv2. Here we use plain SODAR without mask interpolation. R50 and R101 denote ResNet-50 and ResNet-101 backbones. AP$^*$ is mask AP evaluated on the 80 COCO categories subset of LVIS benchmark. Inference speed is measured on V100 GPU. \textit{The inference speed is averaged over the \textbf{val} set}.	\vspace{-1em}}
	\label{main_result}
\end{table*}

\begin{table*}[]
	\centering
	\renewcommand{\tabcolsep}{2.0pt}
	\renewcommand{\arraystretch}{1.1}
	\caption{Results of SODAR with \textbf{mask interpolation}. The ``mask-itp'' means the proposed mask interpolation. SODAR* means SODAR model built on SOLOv2. \textbf{+cls} means increasing classification grid resolution; \textbf{-mask} means reducing mask grid resolution. We also include the results of hybird setting, \textit{i.e.}, increasing classification grid and reducing mask grid (+c-m). \textit{Note the inference speed is averaged over the full \textbf{val} set}.}
	\label{main_result_mask_interpolation}
	\begin{tabular}{c|c|cccccc|ccccccc}
		\toprule
		Methods & mask-itp & AP & AP$_{50}$ & AP$_{75}$ & AP$_{s}$ & AP$_{m}$ & AP$_{l}$ & AP$^*$ & AP$^*_{50}$ & AP$^*_{60}$ & AP$^*_{70}$ & AP$^*_{80}$ & AP$^*_{90}$ & FPS \\\midrule
		SOLO-R101 & -    &37.1  & 58.1  & 39.5  & 15.6  & 41.0  & 55.5  & 38.6  & 59.7  & 53.5  & 45.8  & 34.3  & 15.6  & 11.5 \\
		SODAR-R101 & -  & 38.6  & 58.9  & 41.7  & 16.5  & 42.6  & 57.7  & 41.2  & 61.9  & 55.8  & 48.4  & 37.7  & 18.5  & 10.2 \\ 
		SODAR-R101& +cls  & \textbf{39.1}  & 59.2  & 42.1  & 17.1  & 43.6  & 58.0  & \textbf{41.6}  & 62.6  & 57.0  & 48.7  & 38.0  & 18.6 & 10.0\\
		SODAR-R101& -mask  & 38.8  & 59.4  & 41.8  & 16.5  & 42.8  & 58.3  & 41.1  & 61.5  & 56.0  & 48.4  & 37.6  & 18.7  & \textbf{10.9}\\
		decoupled-SOLO~\cite{wang2019solo}& -   & 37.9  & 58.9  & 40.6  & 16.4  & 42.1  & 56.3  & 39.6  & 60.5  & 54.5  & 46.5  & 35.6  & 16.7 & 10.6 
		\\\midrule 
		SOLOv2-R101& -  & 38.5  & 59.1  & 41.3  & 17.1  & 42.5  & 56.8  & 41.1  & 61.9  & 55.9  & 48.0  & 37.3  & 18.8  & 13.4 \\
		SODAR*-R101 & -  &39.4  & 60.0  & 42.3  & 16.8  & 43.6  & 58.5  & 42.3  & 62.3  & 56.9  & 49.7  & 38.9  & 20.5  & 12.2 \\
		SODAR*-R101 &+cls& \textbf{39.7}  & 60.6  & 43.1  & 17.9  & 43.8  & 58.8  & \textbf{42.8}  & 63.3  & 57.6  & 50.3  & 39.4  & 20.7 & 11.8 \\
		SODAR*-R101&-mask & 39.2  & 60.0  & 42.2  & 16.9  & 43.3  & 58.5  & 42.2  & 62.4  & 56.5  & 49.4  & 38.8  & 20.1 & \textbf{12.7} \\ \midrule 
		SODAR-R101 &+c-m & 39.0  & 59.5  & 42.0  & 16.9  & 42.8  & 58.4  & 41.4  & 61.6  & 56.2  & 48.4  & 37.7  & 18.8  & 10.4\\
		SODAR*-R101 &+c-m& 39.4  & 60.1  & 41.9  & 17.0  & 43.5  & 58.5  & 42.4  & 62.5  & 56.9  & 50.2  & 39.6  & 20.7 & 12.4 \\
		\bottomrule
	\end{tabular}
\end{table*}

\begin{table*}[]
	\centering
	\renewcommand{\tabcolsep}{2.0pt}
	\renewcommand{\arraystretch}{1.1}
	\caption{Results by applying the proposed \textbf{deformable neighbor sampling}. d-neighbor means the deformable neighbor sampling. \textit{Note the inference speed is averaged over the full \textbf{val} set}.}
	\label{main_result_deform_neighbor}
	\begin{tabular}{c|c|c|cccccc|ccccccc}
		\toprule
		Methods & mask-itp & d-neighbor& AP & AP$_{50}$ & AP$_{75}$ & AP$_{s}$ & AP$_{m}$ & AP$_{l}$ & AP$^*$ & AP$^*_{50}$ & AP$^*_{60}$ & AP$^*_{70}$ & AP$^*_{80}$ & AP$^*_{90}$ & FPS \\\midrule
		SODAR-R101& -mask  & - & 38.8  & 59.4  & 41.8  & 16.5  & 42.8  & 58.3  & 41.1  & 61.5  & 56.0  & 48.4  & 37.6  & 18.7  & 10.9\\
		SODAR-R101& -mask  & \checkmark & 39.1  & 59.6  & 41.9  & 16.6  & 43.2  & 58.6  & 41.5  & 62.4  & 57.1  & 48.3  & 37.7  & 18.9  & 10.2\\\midrule
		SODAR-R101& +cls & -  & 39.1  & 59.2  & 42.1  & 17.1  & 43.6  & 58.0  & 41.6  & 62.6  & 57.0  & 48.7  & 38.0  & 18.6 & 10.0\\
		SODAR-R101& +cls & \checkmark  & 39.5  & 59.2  & 42.4  & 17.2  & 43.8  & 59.0  & 42.5  & 63.6  & 56.8  & 49.4  & 39.1  & 20.2 & 9.0\\\midrule
		SODAR*-R101&-mask &-& 39.2  & 60.0  & 42.2  & 16.9  & 43.3  & 58.5  & 42.2  & 62.4  & 56.5  & 49.4  & 38.8  & 20.1 & 12.7 \\ 
		SODAR*-R101&-mask &\checkmark& 39.6  & 60.1  & 42.6  & 17.4  & 43.3  & 59.2  & 42.5  & 62.9  & 56.4  & 49.6  & 39.2  & 20.5 & 11.8 \\ \midrule
		SODAR*-R101 &+cls& - & 39.7  & 60.6  & 43.1  & 17.9  & 43.8  & 58.8  & 42.8  & 63.3  & 57.6  & 50.3  & 39.4  & 20.7 & 11.8 \\
		SODAR*-R101 &+cls& \checkmark  & \textbf{40.0}  & 60.3  & 43.4  & 18.4  & 43.6  & 59.1  & \textbf{43.4}  & 63.6  & 57.8  & 50.3  & 39.9  & 21.3 & 10.7 \\
		\bottomrule
	\end{tabular}
\end{table*}

%


\begin{table*}[h!]
	\centering
	\renewcommand{\tabcolsep}{2.0pt}
	\renewcommand{\arraystretch}{1.1}
	\caption{Model analysis results.  We use the SODAR-R101 model for the experiments. agg-D and agg-S means using the proposed dynamic weight versus alternative static weight for aggregation function. $|\mathcal{N}|$-* means setting different number of neighbors for the aggregation,``-col'' and ``-row'' means column-wise and row-wise neighbors. multi-level-context-* is ablation of the multi-level context information feature. The last three are ablation of the layer number and kernel size, and adding original mask loss.
	}
	\label{model_analysis}
	\begin{tabular}{l|cccccc|cccccc}
		\toprule
		& AP & AP$_{50}$ & AP$_{75}$ & AP$_{s}$ & AP$_{m}$ & AP$_{l}$ & AP$^*$ & AP$^*_{50}$ & AP$^*_{60}$ & AP$^*_{70}$ & AP$^*_{80}$ & AP$^*_{90}$ \\ \midrule
		Baseline & 37.1  & 58.1  & 39.5  & 15.6  & 41.0  & 55.5  & 38.6  & 59.7  & 53.5  & 45.8  & 34.3  & 15.6 \\\midrule
		agg-D & 38.6  & 58.9  & 41.7  & 16.5  & 42.6  & 57.7  & 41.2  & 61.9  & 55.8  & 48.4  & 37.7  & 18.5\\ 
		agg-S & 37.5  & 57.4  & 40.4 & 15.7  & 41.9  & 56.4  & 39.7  & 61.3  & 54.9  & 47.4  & 36.1  & 16.9 \\ \midrule
		no-neighbor & 37.4  & 57.1  & 40.9 & 15.5  & 41.8  & 56.2  & 39.9  & 61.5  & 54.5  & 47.4  & 36.2  & 16.7 \\
		$|\mathcal{N}|$-2-col & 38.0  & 58.3  & 41.1  & 16.0  & 42.0  & 57.5  & 40.6  & 61.6  & 55.1  & 48.1  & 37.6  & 17.9   \\
		$|\mathcal{N}|$-2-row & 38.1  & 58.3  & 41.0  & 16.2  & 42.1  & 57.3  & 40.7  & 61.6  & 55.4  & 47.9  & 37.5 & 17.9   \\
		$|\mathcal{N}|$-4 & 38.6  & 58.9  & 41.7  & 16.5  & 42.6  & 57.7  & 41.2  & 61.9  & 55.8  & 48.4  & 37.7  & 18.5  \\
		$|\mathcal{N}|$-8 & 38.5  & 58.5  & 42.3  & 16.2  & 42.4  & 57.9  & 41.2  & 61.8  & 55.7  & 48.2  & 37.6  & 18.4   \\ \midrule
		multi-level-context-w & 38.6  & 58.9  & 41.7  & 16.5  & 42.6  & 57.7  & 41.2  & 61.9  & 55.8  & 48.4  & 37.7  & 18.5  \\ 
		multi-level-context-w/o & 38.2  & 58.3  & 41.1  & 16.0  & 42.3  & 57.4  & 40.7  & 61.8  & 55.0  & 48.1  & 37.5  & 18.1 \\
		multi-level-context-only & 34.4  & 54.0  & 37.1  & 13.8  & 39.7  & 50.6  & 35.3  & 54.2  & 49.6  & 42.4  & 31.4  & 14.5 \\ \midrule
		conv-layer-1 & 38.1  & 58.2  & 41.2  & 16.1  & 42.1  & 57.5  & 40.7  & 61.1  & 55.6  & 47.8  & 37.1  & 18.1 \\
		conv-layer-2 & 38.4  & 58.8  & 41.4  & 15.7  & 42.4  & 57.6  & 41.0  & 61.3  & 55.3  & 48.0  & 37.1  & 19.3 \\
		conv-layer-3 & 38.6  & 58.9  & 41.7  & 16.5  & 42.6  & 57.7  & 41.2  & 61.9  & 55.8  & 48.4  & 37.7  & 18.5 \\
		conv-layer-4 & 38.4  & 59.0  & 41.0  & 16.1  & 42.3  & 57.9  & 41.0  & 61.5  & 55.5  & 47.7  & 37.0  & 19.2 \\ \midrule
		conv-kernel-1x1 & 38.2  & 57.9  & 41.6  & 16.0  & 42.2  & 57.7  & 40.5  & 61.3  & 55.8  & 47.3  & 37.0  & 18.0 \\
		conv-kernel-3x3 & 38.6  & 58.9  & 41.7  & 16.5  & 42.6  & 57.7  & 41.2  & 61.9  & 55.8  & 48.4  & 37.7  & 18.5  \\ \midrule
		two-stage-loss & 37.2  & 57.1  & 40.3 & 15.5  & 41.2  & 56.7  & 38.8  & 60.4  & 54.5  & 46.9  & 38.9  & 16.5 \\ 
		\bottomrule
	\end{tabular}
\end{table*}

\subsection{Model Analysis}
\label{ablation_study}

\subsubsection{How Does Aggregation Help}
To understand how aggregation helps the model to obtain better segmentation results, we visualize the learned mask representations.
As shown in Fig.~\ref{vis_mask_rep}, the activated area is larger than that of a plain SOLO model whose mask predictions only focus on the target objects.
In addition, we observe they capture complementary context information of neighboring objects which helps to segment the object when combined. 
For example, in Fig.~\ref{vis_mask_rep}, the mask representations at the grid cells covering the monitor object additionally attend to different context information.
In addition to the target monitor instance, the representation ``a'' attends more to the lamp, ``b'' attends to the desk, and ``c'' attends to the people sitting before the monitor.

\subsubsection{Comparison with Direct Mask Prediction Baselines}
As shown in Tab.~\ref{main_result}, SODAR significantly boosts both baselines SOLO and SOLOv2 across different backbones.
For example, with a ResNet-50 backbone, a SOLO model gets 2.1 AP improvement and a SOLOv2 model gets 1.0 AP improvement.
The improvement on ResNet-101 backbone is consistent, \textit{i.e.}, 1.5 AP for SOLO and 0.9 AP for SOLOv2.
Notably, the performance gap becomes larger when evaluating on LVIS dataset (AP$^*$), and especially significant for AP with high IoU (\textit{i.e.}, AP$^*_{70}$, AP$^*_{80}$ and AP$^*_{90}$). 
For instance, the AP of the SOLO-R50 model is improved by 2.1 while the AP$^*$ is improved by 3.0, and the improvement is even higher on AP$^*_{70}$, AP$^*_{80}$ and AP$^*_{90}$, \textit{i.e.}, 3.5, 4.2 and 3.2 for each of them respectively. 
As high IOU AP is a challenging metric that requires much higher segmentation quality, this observation shows effectiveness of the proposed method in improving mask segmentation quality. 
In terms of inference speed, SODAR only introduces small additional cost compared to baselines.
Similarly, we observe a similar trend of AP improvement when augmenting SOLOv2 with our proposed aggregation framework (SODAR$^*$-R101) despite its already high accuracy.


\subsubsection{Incorporating Mask Interpolation}
\label{text_mask_interpolation}
We study two instantiations to demonstrate the merit of mask interpolation. 
1) \textbf{``+cls''}: increasing the classification grid resolution from [40, 36, 24, 16, 12] to [50, 40, 24, 16, 12] for the 5 respective FPN levels while keeping the mask grid resolution unchanged (\textit{i.e.}, [40, 36, 24, 16, 12]); 
2) \textbf{``-mask''}: reducing the mask grid resolution from [40, 36, 24, 16, 12] to [20, 18, 12, 8, 6] while maintaining the grid resolution for classification (\textit{i.e.}, [40, 36, 24, 16, 12]). 
The former aims to improve the discrimination ability for small objects, while the latter aims to reduce the memory and computation cost for generating the mask representations.
Note for the \textbf{``-mask''} scheme, the number of representations is reduced quadratically. For example, grid resolution of 40 corresponds to 1,600 mask representations while only 400 mask representations is needed by halving the resolution to 20.
As shown in Tab.~\ref{main_result_mask_interpolation}, for a SOLO model with ResNet-101 backbone, the \textbf{``+cls''} scheme improves the overall performance from 38.6 to 39.1 in AP. 
The improvement is mainly on small and medium-sized objects as revealed in AP$_{s}$ and AP$_{m}$, \textit{i.e.}, 0.6 AP for both the AP$_{s}$ and AP$_{m}$.
The time overhead is almost negligible (10.0 \textit{vs.} 10.2 FPS).
On the other hand, reducing the mask grid resolution improves the inference speed from 10.2 to 10.9 FPS while giving comparable performance. 
A similar trend is observed for both SOLO and SOLOv2.
Our \textbf{``-mask''} scheme also outperforms decoupled SOLO~\cite{wang2019solo} in terms of both accuracy and speed. We also examine the hybird setting of both increasing classification grid resolution and reducing mask grid resolution. The performance and inference speed are between that of only increasing classification grid resolution and reducing mask grid resolution, \textit{e.g.}, 39.0 AP and 10.4 FPS for SODAR with ResNet-101 backbone. An architecture search may be applied to acquire the best performance-speed trade off setting of classification and mask grid resolution, which is beyond the scope of this work. 

\begin{figure}[H]
	\centering
	\includegraphics[height=0.4\linewidth]{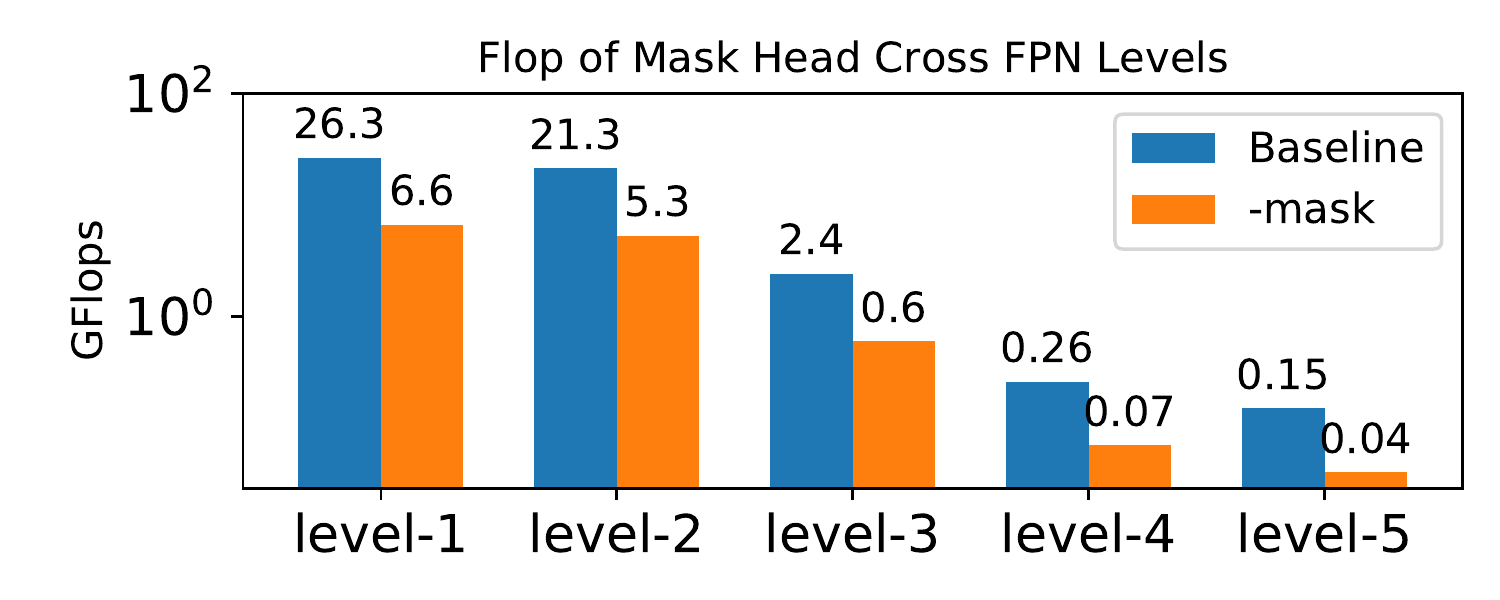}
	\caption{Comparison of mask head FLOPs between baseline SOLO-R101 and SODAR-R101 with mask interpolation of \textbf{``-mask''} scheme. The result is obtained by averaging over COCO {\tt val} set. The FLOPs is reduced roughly 4 times across different FPN levels.}%
	\label{flop_per_level}
\end{figure}

In terms of computation and parameter overhead, as shown in Tab.~\ref{computation_params}, 
incorporating the proposed aggregation method to a ResNet-101-based SOLO model only brings about $\sim$1.4\% (422.8 GFLOPs to 428.3 GFLOPs) increase in the FLOPs and 0.3M increase in parameters.
Adding mask interpolation (\textbf{``+cls''} scheme) and deformable neighbor sampling introduces $\sim$2.3\% (422.8 GFLOPs to 433.8 GFLOPs) and $\sim$3.1\% (422.8 GFLOPs to 435.8 GFLOPs) additional FLOPs, respectively. And the overhead in parameter count is also small, \textit{i.e.}, 0.3M for the \textbf{``+cls''} scheme and 0.6M for further adding deformable sampling.
In contrast, the \textbf{``-mask''} scheme reduces the computation, \textit{e.g.}, adopting the \textbf{``-mask''} scheme for mask interpolation, the computation is reduced from 428.3G (only aggregation) to 390.5G, and the parameter count is reduced from 55.4M to 54.6M.
Fig.~\ref{flop_per_level} shows an analysis that
compares the mask head computation cost of the original SOLO-R101 model and SODAR-R101 with the \textbf{``-mask''} scheme. We observe the FLOPs of mask head is significantly reduced across FPN levels. 

\begin{table}[H]
	\centering
	\renewcommand{\tabcolsep}{2.0pt}
	\renewcommand{\arraystretch}{1.1}
	\caption{Computation and parameter count analysis, with a SOLO model on ResNet-101 backbone. 
	}
	\label{computation_params}
	\begin{tabular}{c|c|c|c|ccccc|ccccccc}
		\toprule
		-&Aggregation&Mask itp.& d-neighbor & FLOPs & Params\\\midrule
		\multirow{5}{*}{\begin{tabular}[c]{@{}l@{}}SOLO\\ -R101\end{tabular}}&  &  &  &422.8G& 55.1M\\
		& $\checkmark$ &  & &428.3G& 55.4M\\
		& $\checkmark$ & +cls & &433.8G& 55.4M\\
		& $\checkmark$ & -mask && 390.5G& 54.6M\\
		& $\checkmark$ & +cls & \checkmark&435.8G& 55.7M\\
		& $\checkmark$ & -mask & \checkmark&391.1G& 54.9M\\
		\bottomrule
	\end{tabular}
\end{table}

\subsubsection{Incorporating Deformable Neighbor Sampling}
We further add the deformable neighbor sampling to the aggregation framework based on the above mask interpolation.
As shown in Tab.~\ref{main_result_deform_neighbor}, with deformable sampling, while the inference speed slightly reduces by 1.0 FPS, we observe an improvement in AP,
\textit{e.g.}, from 39.1 to 39.6 for the SODAR-R101 model and 39.7 to 40.1 for the SODAR*-R101 model with \textbf{``+cls''} scheme.
For the \textbf{``-mask''} scheme, a similar trend is observed,
demonstrating the effectiveness of dynamically adjusting the neighbor sampling locations for aggregation.
The result indicates that it is beneficial to allow the aggregation function to adaptively adjust the neighbor sampling locations, which provides another dimension of freedom to the aggregation.

\subsubsection{Individual Design Choices}
We next carefully examine each proposed component to quantitatively justify our design choices (Tab.~\ref{model_analysis}).
We employ SODAR-R50 which is built on SOLO-R50 to conduct all ablation experiments.
Our findings are summarized as below: 

\textbf{Dynamic Aggregation \textit{vs.} Static Aggregation}.
We compare our dynamic aggregation with a static alternative that uses fixed learned weights for aggregation. As shown in the second and third rows of Tab.~\ref{model_analysis},
the dynamic aggregation scheme outperforms the static counterpart by a large margin, \textit{i.e.}, 37.9 \textit{vs.} 36.5 in AP, and 40.0 \textit{vs.} 38.5 in AP$^*$.
This verifies our assumption that dynamically generated weights are essential to 
enable
adaptive aggregation, and thus contribute to higher performance.

\begin{figure*}[h!]
	\centering
	\includegraphics[width=0.93\linewidth]{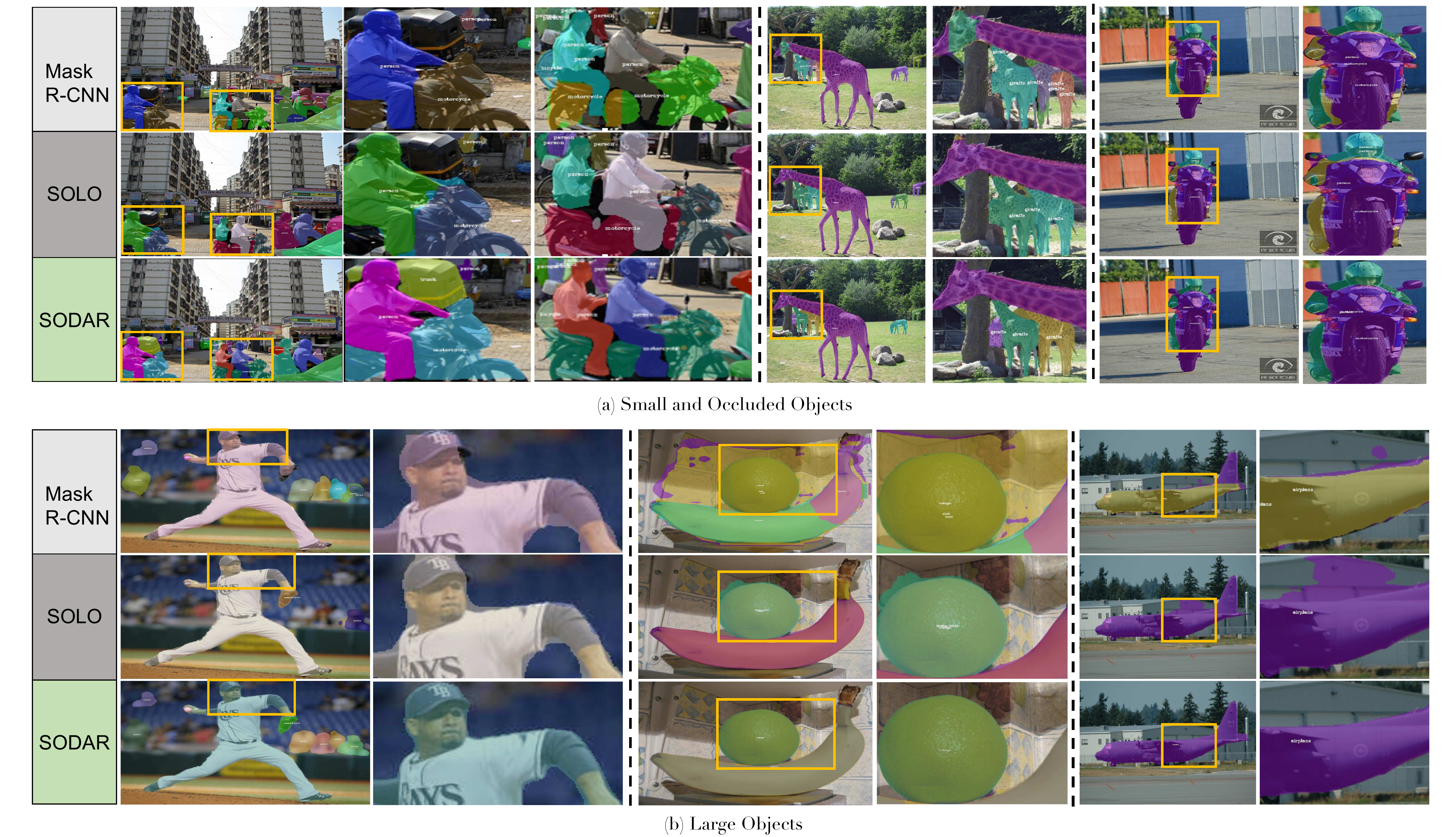}
	\caption{Qualitative comparison of Mask R-CNN, SOLO and SODAR results. Right three columns are zoom-in views of yellow rectangle areas, showing SODAR can handle severe occlusions in crowded scenes (a). It also more accurately segments large objects (b).}
	\label{vis_mask_pred}
\end{figure*}

\begin{table}[h]
	\centering
	\renewcommand{\tabcolsep}{2.0pt}
	\renewcommand{\arraystretch}{1.1}
	\caption{Comparison with existing methods on COCO {\tt test-dev} set, all with ResNet-101 backbone. SODAR* means SODAR model built on SOLOv2. Here we use SODAR with mask interpolation of +cls scheme and deformable neighbor sampling. +grid denotes increasing the grid resolution to [50,40,24,16,12]. We also apply the aggregation method to TensorMask~\cite{chen2019tensormask} and CondInst~\cite{tian2020conditional} (TensorMask+agg, CondInst+agg) $\dagger$ means our re-trained model with the officially released code, note we use implementation for Mask R-CNN~\cite{massa2018maskrcnn} and implementation for Centermask~\cite{centermask2}. \textit{sch.} denotes training schedule, 2x, 3x and 6x means 24, 36 and 72 epochs, respectively. \textit{The inference speed is averaged over the full \textbf{test-dev} set}.}
	\label{test_dev}
	\begin{tabular}{l|cccccccc}
		\toprule
		Methods &  AP & AP$_{50}$ & AP$_{75}$ & AP$_{s}$ & AP$_{m}$ & AP$_{l}$ & FPS & \textit{sch.}\\ \midrule
		MR-CNN$\dagger$~\cite{he2017mask} & 38.2 & 61.3  & 40.9  & 19.5  & 40.4  & 53.3 & 9.1 & 6x\\
		MaskLab+~\cite{chen2018masklab} &  37.3 & 59.8 & 39.6 & 16.9 & 39.9 & 53.5 & - & -\\
		YOLACT~\cite{bolya2019yolact}&  31.2 & 50.6  & 32.8  & 12.1  & 33.3  & 47.1 & 23.4 & 4x\\
		PolarMask~\cite{xie2019polarmask} &  32.1 & 53.7  & 33.1  & 14.7  & 33.8  & 45.3 & 12.3 & 2x\\
		MEInst~\cite{zhang2020mask} & 33.9& 56.2 &35.4 &19.8 &36.1 & 42.3 & - & 3x\\
		TensorMask~\cite{chen2019tensormask} & 37.1 & 59.3  & 39.4  & 17.4  & 39.2  & 51.6 & 2.6 & 6x\\
		BlendMask~\cite{chen2020blendmask} & 38.4 &  60.7 & 41.3 & 18.2 & 41.5 & 53.3 & 11.0 & 3x\\
		CondInst$\dagger$~\cite{tian2020conditional} & 39.3 & 60.6 & 42.3 & 21.2 & 42.1 & 51.8 & 10.3 & 6x\\
		CenterMask$\dagger$~\cite{lee2019centermask} & 39.9 & 59.7 & 43.7 & \textbf{21.7} &42.9 & 51.8 & 14.8 & 6x\\
		SOLO~\cite{wang2019solo}  & 37.8 & 59.5  & 40.4  & 16.2  & 40.6  & 54.2 & 11.6 & 6x\\
		SOLOv2~\cite{wang2020solov2} & 39.7 & 60.7  & 42.9  & 17.1  & 42.9  & 57.5 & 13.6 & 6x\\
		SOLO+grid  & 38.0 & 59.3  & 40.8  & 16.5  & 40.7  & 54.0 & 10.1 & 6x\\
		SOLOv2+grid & 39.8 & 60.4  & 43.0  & 17.1  & 43.0  & 57.4 & 13.1 & 6x\\
		\midrule
		TensorMask+agg & 38.7 & 60.5  & 40.8  & 18.0  & 40.1  & 53.4 & 2.4 & 6x\\
		CondInst+agg & 40.4 & 61.2 & 43.1 & 21.6 & 43.6 & 53.0 & 9.2 & 6x\\
		SODAR (\textbf{ours})&  40.0  & 60.4  & 43.5  & 17.7  & 43.5  & 57.4 & 10.6 & 6x\\
		SODAR* (\textbf{ours})&\textbf{41.0} & \textbf{61.6} & \textbf{44.4} & 18.4  & \textbf{44.4}  & \textbf{59.0} & 12.7 & 6x \\
		\bottomrule
	\end{tabular}
\end{table}

\textbf{Number of Neighbors}.
With \textbf{``no-neighbor''} setting, the model simply reduces to learning a dynamic transformation from the location-aware mask representation into the final mask prediction.
This setting already brings 0.5 AP improvement upon the baseline (\textit{i.e.}, 36.3 vs 35.8). 
When introducing 2 neighboring mask representations ({\bf ``$|\mathcal{N}|$-2-col''} and {\bf ``$|\mathcal{N}|$-2-row}''), the performance is significantly improved. 
Specifically, the performance is improved by about 0.9 AP for the two settings, \textit{e.g.}, from 36.3 to 37.3 for column-wise neighbor (upper and lower grid cells) and 37.2 for row-wise neighbor (left and right grid cells). 
Including both row and column neighbors further improves the performance to 37.9 AP. 
This observation indicates the two orthogonal spatial dimensions encode different context information and improve one another when combined.
However, the gain in AP is diminishing when $|\mathcal{N}|=8$.
This is possibly because the representation from far-away grid cells contain context that is irrelevant to the current cell, especially for small-sized objects. 

\textbf{Multi-level Context Information}.
Next we examine the effect of multi-level context information on the proposed dynamic aggregation module.
When the multi-level context feature is removed from the aggregation process, the performance drops from 37.9 to 37.1 in AP and 40.0 to 39.2 in AP$^*$, demonstrating multi-level feature is beneficial for supplementing dynamic aggregation with bottom-up semantic information. On the other hand,
if we remove mask representations and only use context information to generate the final mask segmentation, AP drops steeply to 33.8. 
This means the local mask representation is crucial for segmenting the objects in the proposed aggregation module while the context information is complementary.

\textbf{Number and Size of Dynamic Convolution Kernels}.
For the dynamic aggregation network, the best performance is achieved with 3 convolution layers. That is, with 3 layers, the performance is improved to 37.9 AP, compared to 37.1 with 1 layer and 37.6 with 2 layers.
Beyond that, the improvement is diminishing, \textit{i.e.}, 4 layers only yielding 37.8 AP.
For the convolution kernel size, 3$\times$3 is better than 1$\times$1 (37.9 vs 37.7 in AP), meaning the aggregation layers needs larger context to achieve higher performance.

\textbf{Supervision of Mask Representations}.
As described earlier, our SODAR implicitly supervises the learning of intermediate mask representations by back-propagating via the aggregation function. Here, we also compare with an alternative solution that employs loss supervision for the mask representations as done in ~\cite{wang2019solo} and \cite{wang2020solov2}.
In this way, the model is supervised to learn both the aggregation function and per grid cell object mask prediction.
Training in this manner converts the SODAR into a two-stage refinement framework, which first predicts a set of location-aware masks, followed by a mask refinement with the designed neighbor aggregation.
Although this alternative ({\bf ``two-stage-loss''}) performs better than the baseline (\textit{i.e.}, 35.8 vs 36.4 in AP and 37.0 vs 38.1 in AP$^*$),
its performance is much worse when compared with the unconstrained counterpart (37.9 vs 36.4 in AP and 40.0 vs 38.1 in AP$^*$). 
This shows the importance of unconstrained local mask representation learning, which enables each local representation to capture useful context information beyond the instance itself, and thus generates higher quality masks when aggregated. 

\begin{figure*}[h!]
	\centering
	\includegraphics[width=0.93\linewidth]{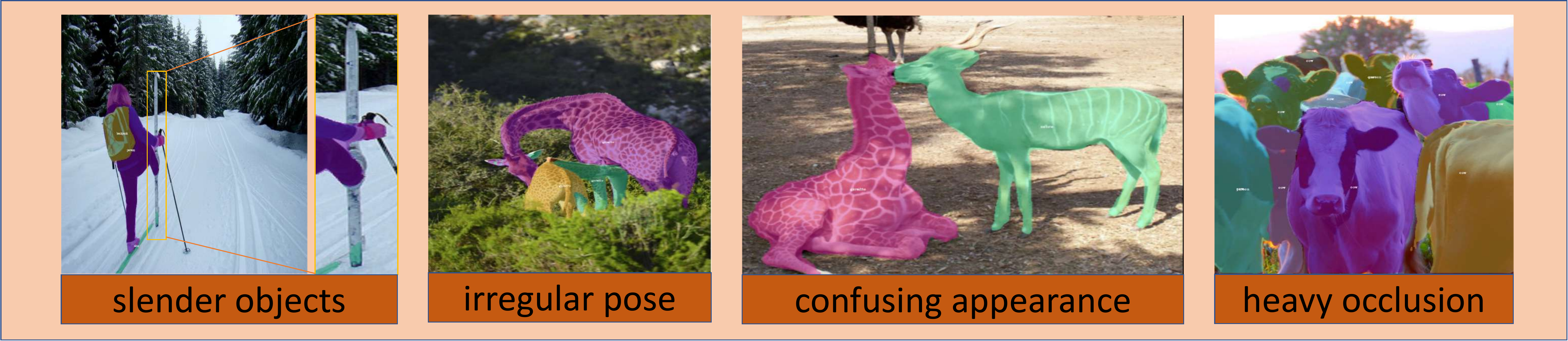}
	\caption{Example failure results on the COCO \textit{test-dev} set, with a ResNet-101 backbone. The example cases are: 1) slender object: the model cannot well segment the ski. 2) irregular pose: the small giraffe is with a irregular pose, causing a false positive prediction. 3) confusing appearance: the goat with zebra alike stripes on its body is mis-recognized as a zebra. 4) heavy occlusion: the herd of cows are difficult to segment correctly due to the heavy occlusion.}
	\label{failure_case}
\end{figure*}

\subsection{Results on COCO test-dev}
\subsubsection{Comparison with Existing Methods on COCO test-dev Set}
We compare our method with existing instance segmentation methods: Mask R-CNN~\cite{he2017mask}, PolarMask~\cite{xie2019polarmask}, YOLACT~\cite{bolya2019yolact}, TensorMask~\cite{chen2019tensormask}, CenterMask~\cite{xie2019polarmask}, BlendMask~\cite{chen2020blendmask}, SOLO~\cite{wang2019solo} and SOLOv2~\cite{wang2020solov2} on the COCO {\tt test-dev} split. 
As shown in Tab.~\ref{test_dev}, our SODAR models built on SOLO and SOLOv2 outperform all baseline methods with ResNet-101 backbone.
All sub AP terms except for AP$_{s}$ are better than baseline methods. 
By applying dynamic convolution~\cite{wang2020solov2} to predict the mask representations as in SOLOv2, SODAR achieves 41.0 mask AP. 
We also test our method on other dense one-stage instance segmentation methods, \textit{e.g.}, TensorMask~\cite{chen2019tensormask} and CondInst~\cite{tian2020conditional}. We incorporate the aggregation module into those two models and observed consistent improvement, \textit{i.e.}, 1.6 AP for TensorMask and 1.1 AP for CondInst. Specifically, for TensorMask, we first remap the sliding window segmentation mask back to the original image space, and then perform the proposed aggregation; for CondInst, the implementation is similar to that of SOLOv2. For the two experiment, we did not adopt the ``+cls" scheme as they do not adopt fixed classification grid as in SOLO and SOLOv2.

\subsubsection{Qualitative Results}
We also present some qualitative results in Fig.~\ref{vis_mask_pred} to illustrate the superiority of proposed aggregation.
SODAR better handles occlusion and has higher quality segmentation masks for both small and large objects, \textit{e.g.}, in the first example of Fig.~\ref{vis_mask_pred} (a), Mask R-CNN and SOLO cannot well segment the the motorcycle ridding by the two people due to occlusion, while SODAR can segment it more precisely. And in Fig.~\ref{vis_mask_pred} (b), SODAR predicts higher quality masks that better delineate the shape of large objects. 
As shown in Fig.~\ref{failure_case} are some common failure case examples of SODAR due to irregular object shape and pose, confusing appearance and heavy occlusion, which have been open problems in the computer vision community and we leave those as our future work.

\section{Conclusion}
In this work, we identify the discarded neighboring mask prediction issue based on the recent state-of-the-art one-stage instance segmentation model. Based on the observation we develop a novel aggregated learning framework that leverages rich neighboring predictions to improve the final segmentation quality. 
The framework in turn enables the network to generate interpretable location-aware mask representations that encodes context objects information. 
We further introduce several model designs that enable us to build a new instance segmentation model named SODAR.
Extensive experimental results show our simple method significantly improves the instance segmentation performance of baseline models. We believe our findings provide useful insights for future research on instance segmentation.

%
%


\ifCLASSOPTIONcaptionsoff
\newpage
\fi




\bibliography{arxiv}
\bibliographystyle{IEEEtran}
\end{document}